# Learning Combined Set Covering and Traveling Salesman Problem

Yuwen Yang*, Jayant Rajgopal

Department of Industrial Engineering

University of Pittsburgh

Pittsburgh, PA 15261

*Corresponding author:

1032 Benedum Hall, Department of Industrial Engineering, University of Pittsburgh, Pittsburgh, PA 15261

yuwen.yang@pitt.edu

# Learning Combined Set Covering and Traveling Salesman Problem

## Abstract

The Traveling Salesman Problem is one of the most intensively studied combinatorial optimization problems due both to its range of real-world applications and its computational complexity. When combined with the Set Covering Problem, it raises even more issues related to tractability and scalability. We study a combined Set Covering and Traveling Salesman problem and provide a mixed integer programming formulation to solve the problem. Motivated by applications where the optimal policy needs to be updated on a regular basis and repetitively solving this via MIP can be computationally expensive, we propose a machine learning approach to effectively deal with this problem by providing an opportunity to learn from historical optimal solutions that are derived from the MIP formulation. We also present a case study using the World Health Organization's vaccine distribution chain, and provide numerical results with data derived from four countries in sub-Saharan Africa.

# 1. Introduction and literature review

The Traveling Salesman Problem (TSP) is among the most intensively studied combinatorial optimization problems by both the operations research and the machine learning communities. It has been widely applied in areas such as planning, manufacturing, genetics, neuroscience, telecommunication, healthcare, supply chains and logistics [1]. The objective in TSP is to find the shortest route that visits each of a given set of locations and returns to the origin. We consider a TSP in combination with the Set Covering Problem (SCP), where the goal of the SCP in this context is to select an optimal subset of locations for facilities, so as to "cover" all locations in the original set of locations [2], [3]. SCP is also a well-studied problem in the operations research community and has been broadly applied in areas such as facility planning, healthcare, and supply chains [4].

TSP by itself is a hard problem. Exact algorithms such as branch-and-bound, cutting-plane and branch-and-cut methods often slow down when the number of nodes exceeds a few hundred [1]. Large-scale TSP is thus mainly solved by approximation algorithms and heuristics to find high quality solutions that are within 2–3% (say) of the optimum [5]–[8]. Some of the widely used TSP approximation algorithms and heuristics include Christofides' algorithm [9], 2-opt moves [10], 3-opt moves [11], [12], the Lin–Kernighan (LK) method [13], [14], large-step Markov chains [15], [16], stem-and-cycle method [17], [18], and ant colony optimization [19]. Interested readers can refer to [1], [5], [20], [21] for more details about TSP approximation algorithms and heuristics.

When combined with SCP, TSP takes on even more complexity. With a given set of facility locations, TSP only considers transportation costs and its binary decision variables only determine whether or not location $i$ is followed by location $j$. When combined with SCP, the problem first needs to identify locations at which to open facilities, and then come up with a route that visits each such location before returning to the origin. In doing this it needs to consider an overall cost that is the sum of facility costs, customer assignment costs, and transportation costs. The binary decision variables determine whether or not a facility is open at location $i$ is open, whether a location $j$ is assigned to an open facility at location $i$, and whether location $k$ is followed by location $l$ in the TSP route. Clearly, the combined SCP and TSP problem raises more issues related to tractability and scalability than either SCP or TSP separately.

In Section 2 we formulate the combined Set Covering and Traveling Salesman Problem as a mixed integer program (MIP). Solving this MIP can theoretically provide us with the optimal solution, but the run time explodes exponentially as the problem size increases. In addition, every time we re-solve the MIP model using new inputs and parameters, it typically starts from scratch and there is often no obvious method to incorporate information from historical solutions. Therefore, in any application where the above combinatorial problem needs to be solved repeatedly over time with different input values, an approach that relies on MIP can become computationally expensive. For example, in an application that we describe in Section 4, the problem needs to be solved and the solution needs to be updated every month over hundreds of locations.

If a training dataset can be derived from a set of historically solved problems, machine learning (ML) could be a natural candidate to effectively deal with this repeated combinatorial situation. The proposed work aims at providing a tractable and scalable learning-based approach to solve the combined SCP and TSP problems. We first formulate this problem as a mixed integer programming model and discuss relevant issues in Section 2. We then propose a learning-based mechanism to efficiently deal with this problem in Section 3. In Section 4, we provide an illustration of our approach with an application to mobile clinics and vaccination outreach that arises in the context of the World Health Organization's (WHO) Expanded Programme on Immunization (EPI) , and present numerical results using data derived from the WHO and four countries in sub-Saharan Africa. Finally, Section 5 discusses some related issues and Section 6 summarizes this paper.

Several recent studies have been conducted by both the operations research and the machine learning communities to try and incorporate machine learning into combinatorial optimization problems. Many of these studies aim at end-to-end learning methods that train the ML model from discrete optimization problems and directly output solutions from the input instance. For instance, Vinyals et al. introduced the pointer network [22] as a recurrent neural network (RNN) that sequentially takes all the nodes in the graph as input and outputs the TSP route using a mechanism that is similar to the graph attention mechanism [23] that is normally used to focus only on a subset of the input. Using a similar model, Bello et al. trained a reinforcement learning model and defined its reward signals as tour lengths [24]. Instead of using RNN to process the input, Kool and Welling utilized graph neural networks (GNN) [25] after adding attention to establish a similar model [26]. GNN can also be derived to learn the node selection policy [27]. Using a different approach, Nowak et al. approximated a double stochastic matrix [28] and Emami and Ranka used Sinkhorn Policy Gradient [29] in the GNN output in order to characterize the permutation. Kaempfer and Wolf developed a Multiple Traveling Salesmen Problem solver using Permutation Invariant Pooling Networks [30]. Bronstein et al. overviewed geometric deep learning problems with several applications, solutions, difficulties, and future directions [31].

Machine learning can also be used to retrieve meaningful properties of optimization problems, or even alongside the optimization as part of the optimization algorithm. This class of approaches include learning to branch [32]–[37], learning to cut [38], [39], learning when to use Dantzig-Wolf decomposition [40], learning how to disaggregate the problem [41], learning where to linearize a mixed integer quadratic problem [42], learning tactical solutions under imperfect information [43], and learning as a modeling tool [44]. These studies, including those end-to-end learning approaches, often face feasibility, modeling, scaling, and data generation challenges, and are still mostly in the exploratory stages [45].

On the other hand, deep learning, as a sub-field of machine learning, has advanced dramatically with the growth of large datasets and computational power, and has led to breakthroughs on a variety of tasks including speech recognition, machine translation, objective detection, and computer vision [46]. Deep learning often outperforms other learning algorithms when exploring high dimensional spaces and large datasets [47], [48]. Many researchers have been using deep learning to solve combinatorial optimization problems. In fact, many of the approaches mentioned above belong to this category; these include Pointer Networks [22], [24], GNN [23], [25]–[28], Sinkhorn Policy Gradient [29], Permutation Invariant Pooling Networks [30], Geometric learning [31], and others [38]–[40], [43]. This paper aims at providing one of the early approaches for an end-to-end learning algorithm for a particular combinatorial optimization problem via deep learning.

## 2. MIP formulation

The Combined Set Covering and Traveling Salesmen Problem can be formulated using the following mathematical programming model:

*Parameters*

$n$: Total number of targeted customer locations

$i$: Index of locations; $1 \leq i \leq n$ for targeted customer locations; $i = 0, n+1$ for the origin

$f_i$: Fixed cost of running a facility at location $i$

$c_{ij}$: Variable cost of assigning a location $j$ to a facility at location $i$; $c_{ii} = 0$

$p$: Average transportation cost per unit of distance

$d_{ij}$: Distance between location $i$ and location $j$ (with $d_{ii} = 0$ and $d_{ij} = d_{ji}$)

$D$: Maximal coverage distance (MCD)

$a_{ij} \in \{0,1\}$: 1 if location $j$ is within a distance $D$ from location $i$

*Variables*

$X_{ij} \in \{0,1\}$: 1 if location $j$ is assigned to facility at location $i$, 0 otherwise;

$Y_i \in \{0,1\}$: 1 if there is a facility at location $i$, 0 otherwise

$Z_{ij} \in \{0,1\}$: 1 if location $i$ is followed by location $j$ on the trip route

$U_i$: Cumulative number of stops visited during a trip when arriving at facility location $i$

**Program 1**:

$$Min \sum_{1 \leq i \leq n} f_i Y_i + \sum_i \sum_{1 \leq j \leq n} c_{ij} X_{ij} + \sum_i \sum_j p d_{ij} Z_{ij} \tag{1}$$

*subject to*

$$X_{ij} \leq a_{ij} \qquad \text{for } \forall i,j \tag{2}$$

$$X_{ij} \leq Y_i \qquad \text{for } \forall 0 \leq i \leq n, 1 \leq j \leq n \tag{3}$$

$$\sum_{i \leq n} X_{ij} = 1 \qquad \text{for } \forall 1 \leq j \leq n \tag{4}$$

$$\sum_j Z_{0j} = 1 \tag{5}$$

$$\sum_j Z_{j0} = 0 \tag{6}$$

$$\sum_i Z_{i(n+1)} = 1 \tag{7}$$

$$\sum_i Z_{(n+1)i} = 0 \tag{8}$$

$$\sum_j Z_{ij} = \sum_j Z_{ji} \qquad for\ \forall 1 \leq i \leq n, \tag{9}$$

$$\sum_j Z_{ij} = Y_i \qquad for\ \forall 1 \leq i \leq n \tag{10}$$

$$U_0 = 0 \tag{11}$$

$$1 \leq U_i \leq \sum_i Y_i \qquad for\ \forall i \geq 1 \tag{12}$$

$$U_i - U_j + MZ_{ij} \leq M - Y_j \qquad for\ \forall i,j \tag{13}$$

$$Z_{ii} = 0 \qquad for\ \forall i \tag{14}$$

$$X_{ij} \in \{0,1\} \qquad for\ \forall i,j \tag{15}$$

$$Y_i \in \{0,1\} \qquad for\ \forall i \tag{16}$$

$$Z_{ij} \in \{0,1\} \qquad for\ \forall i,j \tag{17}$$

$$U_i \geq 0 \qquad for\ \forall i \tag{18}$$

Note that the objective function (1) minimizes the sum of the facility location, assignment, and transportation costs. Constraint set (2) ensures that a location can only be assigned to a facility that is within the maximal coverage distance (MCD), in which case $a_{ij} = 1$, and 0 otherwise. Constraint set (3) ensures that the assignment is to an existing facility. Constraint set (4) ensures that each location is assigned to exactly one facility. These three sets of constraints define a typical Set Covering Problem.

The next few sets of constraints relate to the Traveling Salesman Problem. Note that node 0 denotes the beginning node of a trip and node $(n+1)$ is the final node of a trip; both represent the origin. Constraint sets (5) and (6) imply that the trip departs from the origin (0) exactly once, while Constraint sets (7) and (8) imply that the trip enters back into the origin ($n+1$) exactly once. Constraint set (9) ensures that the flow that enters and departs any location $i$ is balanced. Constraints (5) – (9) thus ensure that the trip is indeed a (0)-($n+1$) path.

Constraint set (10) states that the trip enters and departs a location $i$ exactly once if there is a facility at this location (i.e., $Y_i = 1$). Constraint set (11) – (13) is the MTZ subtour elimination constraints introduced by Miller, Tucker, and Zemlin [49]. Note that if $j$ follows $i$ in the TSP route, $Z_{ij} = 1$ implies $Y_j = 1$ and $U_j \geq U_i + Y_j > U_i$. Suppose there exists a subtour ($i, j, \ldots i$), with $i \neq 0, n + 1$. Then, $U_j > U_i > U_j$ leads to a contradiction. Note that if location $i$ is not a part of the trip (in which case $Y_i = 0$ by Constraint set (10)), the values of $U_i$ are irrelevant to the problem as long as they satisfy the constraints. Constraint sets (14) – (18) are self-explanatory.

Having to solve this MIP on a regular basis can be computationally expensive or even impossible when *n* is large. In the next section, we introduce a machine learning approach that leverages the MIP to solve this problem.

## 3. Proposed learning mechanism

We start by formally defining the learning models to solve the combined SCP and TSP, and discuss how we iteratively update the model parameters. Suppose the optimal solution to **Program 1** is given by the vector $(X^*, Y^*, Z^*)$. The proposed framework is to establish two supervised machine learning models that can be trained to find this vector given a set of problem parameters. In particular, given a graph $G$ along with its associated location and cost information, **Model 1** will be used to derive $(X^*, Y^*)$. That is, Model 1 is trained to provide the portion of the optimal solution corresponding to the set covering problem, while also accounting for the TSP cost that this results in (in addition to the location and assignment costs). Recall that $Y_j$, as a component of $Y$, is a binary variable that represents whether the facility in a particular location $j$ is open or not. Model 1 is trained to generate each component $Y_j$ of the vector $Y$ as a value between 0 and 1 that corresponds to the probability that in the optimum solution, the facility at location $j$ is open. Denoting this vector of probabilities by $g(Y)$ we could define **Model 1** via the following map:

$$\text{Model 1 (SCP predictor): } G \rightarrow g(Y)$$

Note that $g(Y)$ is used to derive a vector $\hat{Y}$ that is a prediction of $Y$, and also a vector $\hat{X}$ that is a prediction of assignments given by $X$. We will discuss this later in Section 3.4.

On the other hand, **Model 2** establishes the relationship between $(G, Y)$ and $(Z)$, i.e., it is trained to use the graph described by $G$ and the facility locations defined by $Y^*$ to determine the optimal TSP sequence in $Z^*$. In particular, given $Y$, Model 2 is trained to generate a set of values between 0 and 1 for each $Z_{ij}$ that correspond to the probability that location $i$ is followed by location $j$ in the optimal sequence. Defining these probabilities by matrix $p(Z)$, we could define Model 2 via the following map:

$$\text{Model 2 (TSP predictor): } (G, Y) \rightarrow p(Z)$$

In the following sections we provide the detailed methodology to train each of **Model 1** and **Model 2**, and discuss how the models can be used to derive a combined solution that is guaranteed to be feasible.

### 3.1 Data generation and labeling

To begin with, we generate a data set **Train** with graph information $G$ that contains candidate facility locations and all relevant cost information. We also generate an example data set **Test** with graph information $G'$. For evaluation purposes, the data sets $G'$ and $G$ should be drawn from the same overall pool, i.e., the elements of these two data sets, although different, resemble each other in some sense. Finally, we generate another data set **Test$_{new}$** with graph information $G^{new}$ that represents significantly different problems that are unseen in data sets **Train** and **Test**.

We then use the MIP formulation that is proposed in Section 2 to solve the optimization problem on each of the instances in $G$, $G'$, and $G^{new}$. For each instance, we use the optimal solution $(X^*, Y^*, Z^*)$ to generate the corresponding vectors $g(Y^*)$ and $p(Z^*)$; note that for these instances $g(Y^*)$ and $p(Z^*)$ have values of 0 or 1 for all components since the "probabilities" from our optimal solutions are either 0 or 1. We also document each of the individual cost component in the optimal solution, namely, the optimal facility location cost, assignment cost, and TSP cost.

## 3.2 Training machine learning Model 1 (SCP predictor)

In this step, we establish a Neural Network on **Train** that maps $G$ to $g(Y)$, where the $j^{th}$ element of $g(Y)$ is the probability that $Y_j^* = 1$. From a pure optimization perspective, we would ideally like this value to be 1 or 0, depending on whether the facility at $j$ is open or closed. Our model is trained via back propagation, and the hyperparameters in the learning process, such as the network architecture, the number of hidden layers and hidden units, the learning rate, the mini-batch size, activation functions and the number of epochs are tuned iteratively. To accomplish this, we further split dataset **Train** into a training set that comprises 90% of the elements in **Train** and a validation set that comprises the remaining 10% of the data set. Within an iteration, the hyperparameters are fixed and the model is trained using the training set with a pre-set number of epochs (one of the hyperparameters). At the end of each iteration, we obtain a model corresponding to the current set of hyperparameters. We then evaluate this model on the validation set. The hyperparameters are then altered for the next iteration based on how the model performs on the validation set. If overfitting is observed, i.e., the model performs significantly better on the training set and worse on the validation set, then regularization layers (typically Dropout layers) are added, and the number of activations and layers are reduced. On the other hand, if underfitting is observed, i.e., the model performs poorly on both sets, then more layers and activations are added. This process is repeated until no significant improvements are observed.

It is important to note that in general, there is no guarantee that the integer values obtained by directly rounding the fractional $g(Y)$ will lead to the optimal solution of **Program 1**. In fact, these integer values might not even lead to feasible solution. In Section 3.4 we will present a detailed discussion on how to convert $g(Y)$ into a feasible Combined SCP and TSP solution.

### 3.3 Training machine learning Model 2 (TSP predictor)

We train another Neural Network on **Train** that maps $(G, Y)$ to $p(Z)$, where element $(i, j)$ of $p(Z)$ is the probability that $Z_{ij}^* = 1$. That is, it uses the graph information described by $G$ and the facility locations defined by $Y$ to generate a matrix of probabilities $p(Z)$ corresponding to the optimal TSP solution. The training process for **Model 2** also follows the same approach described in Section 3.2 for **Model 1** of dividing **Train** into a training and validation set with similar hyperparameter tuning over a predetermined number of epochs. In the training process, we use the optimal vector $Y^*$ with its associated TSP path since the objective here is to train this model to find the optimal path for a given set of locations on a given graph. In the next section we discuss how to utilize **Model 2** to generate a TSP route from $p(Z)$ that visits each open location.

### 3.4 Obtaining a feasible solution to Program 1

In Section 3.2 and Section 3.3, we developed probability vectors $g(Y)$ and probability matrix $p(Z)$. In this section, we use these results to derive a feasible solution to the combined SCP/TSP.

Recall that $g(Y)$ corresponds to the probabilities associated with opening a facility at each location. For each example, we define a threshold $\alpha$ and let $\hat{Y}_i = 1$ if $g(Y_i) \geq \alpha$ and $\hat{Y}_j = 0$ for all $j$ with $g(Y_i) < \alpha$, i.e., we use $\alpha$ to convert the vector of probabilities $g(Y)$ into the discrete estimator $\hat{Y}$ for use as the SCP solution. Note that in order to ensure that $\hat{Y}$ is feasible, we need to ensure that (i) every location $i$ either has an open facility, or (ii) is assigned to a location $j$ with an open facility that is within a distance $D$ from it, i.e., $\hat{Y}_j = 1$ with $a_{ij} = 1$. We use traversal search in **Algorithm 1** to go over all examples and locations to check if either of these two conditions is meet and force any location $i$ that does not meet either to have an open facility. Note that all locations with open facilities are obviously assigned to the facility there, i.e., $\hat{X}_{ii} = 1$ if $\hat{Y}_i = 1$, and any location where there is no open facility is assigned to the feasible open facility location that leads to the smallest assignment cost. The algorithm is summarized in Table 1.

**Table 1. Algorithm 1** to obtain SCP solution

---
**Input**:
Optimization parameters $a_{ij}^{exp}, d_{ij}^{exp}, c_{ij}^{exp}, f_i^{exp}$ for each example $exp$ and location $i, j$
Machine learning **Model 1** result $g^{exp}(Y_i)$ for each example $exp$ and location $i$
Parameter $\alpha$

---
**for** $exp$ in index of $G$, $G'$, or $G^{new}$:
    **for** $i$ in Index of locations:
        let $\hat{X}_{ij}^{exp} = 0$ for all $j$ in index of locations
        **if** $g^{exp}(Y_i) < \alpha$:
            **if** $\exists\ K$ such that $\forall\ k \in K$, $a_{ij}^{exp} = 1$ and $g^{exp}(Y_i) > \alpha$:
                let $\hat{Y}_i^{exp} = 0$
                let $\hat{X}_{ji}^{exp} = 1$ where $j = \min_{l \in L} l$, and $L = \operatorname*{argmin}_{k \in K} c_{ki}^{exp}$
            **else**:

$$\text{let } \hat{Y}_i^{exp} = 1$$
$$\text{let } \hat{X}_{ii}^{exp} = 1$$

    **else**:
$$\text{let } \hat{Y}_i^{exp} = 1$$
$$\text{let } \hat{X}_{ii}^{exp} = 1$$

**Output**:

SCP solution $\hat{X}$ and $\hat{Y}$ with $\hat{X}_{ij}^{exp}, \hat{Y}_i^{exp}$ for each example $exp$ and location $i, j$

**Proposition 1:** $\hat{X}$ and $\hat{Y}$ that are generated by **Algorithm 1** satisfy Constraint sets (2) – (4) in **Program 1**.

**Proof:** The outer *for* loop iterates through each of all examples in our **Train**, **Test**, and **Test**<sub>new</sub> dataset, which corresponds to an optimization problem. The inner *for* loop further iterates all facility location $i$, whose $g^{exp}(Y_i)$ should be either $< \alpha$ or $\geq \alpha$ that are represent by the outer if/else conditional statement. In the first part of the inner *if/else* statement, Constraint set (2) and (3) are satisfied by the condition, i.e., $X_{ji}^{exp} = 1$ only if $a_{ij}^{exp} = 1$ and $g^{exp}(Y_i) > \alpha$ (which leads to $\hat{Y}_i^{exp} = 1$ in the second part of the outer if/else conditional statement). In all other parts, Constraint set (2) and (3) are also satisfied by ensuring that $\hat{X}_{ii}^{exp} = 1$ with $\hat{Y}_i^{exp} = 1$ and $a_{ii}^{exp} = 1$. Additionally, throughout the inner and outer if/else conditional statement, there exists one and only one $j$ such that $X_{ji}^{exp} = 1$. Note that $j$ is the location with smallest assignment cost (and with the smallest index if there are multiple such $j$), or $j = i$.

So far we have obtained $\hat{X}$ and $\hat{Y}$ that constitute a feasible solution to the SCP. Next, we input $(G, \hat{Y})$ ($G'$ or $G^{new}$ in **Test** and **Test**<sub>new</sub> respectively) into **Model 2** to obtain $p(Z)$. Note that we do not input the true optimal solution $Y^*$ into Model 2, because $Y^*$ is assumed to be unknown when conducting a valid end-to-end evaluation and comparison. In order to derive from $p(Z)$ a feasible set of TSP routes given by $\widehat{Z}$, the routes are required to visit each of the open location in $\widehat{Y}$ once, and only once. Note that the routes have to start from the origin (indexed by 0) and return to the origin (indexed by $n+1$). We start the transformation process with node 0, and look at $S^{exp}$, the set of open facilities in an instance $exp$ with $\{i: \widehat{Y}_i^{exp} = 1$ for all $i \leq n\}$. We find the node $j$ in $S^{exp}$ with the largest probability $\hat{p}^{exp}(Z_{0j})$ (and with the smallest index if there are multiple such $j$), i.e., $j$ is the most promising stop to follow node 0 in the optimal TSP route. We let $\hat{Z}_{0j} = 1$ and remove $j$ from $S^{exp}$. We continue with a similar process while adding open facilities to the TSP route until $S^{exp}$ is empty. We summarize this algorithm in Table 2.

**Table 2**. **Algorithm 2** to obtain TSP solution

| |
|---|
| **Input**: |
| Machine learning **Model 2** result $\hat{p}^{exp}(Z_{ij})$ for each example $exp$ and locations $i, j$ |
| Set of open facility $S^{exp} = \{i: \widehat{Y}_i^{exp} = 1$ for all $i \leq n\}$ |
| **for** $exp$ **in** index of $G$, $G'$, and $G^{new}$: |
|     let $\hat{Z}_{ij}^{exp} = 0$ for all $i$ and $j$ in index of locations |
|     let $i = 0$ |
|     let $S^{exp} = S^{exp} - \{i\}$ |
|     **while** $S^{exp} \neq \phi$: |
|         let $\hat{Z}_{ij}^{exp} = 1$ where $j = \min_{l \in L} l$, and $L = \underset{k \in S^{exp}}{\mathrm{argmax}}\, \hat{p}^{exp}(Z_{ik})$ |
|         let $S^{exp} = S^{exp} - \{j\}$ |
|         let $i = j$ |
|     let $\hat{Z}_{i(n+1)}^{exp} = 1$ |
| **Output**: |
| TSP solution $\widehat{Z}$ with $\hat{Z}_{ij}^{exp}$ for each example $exp$ and location $i, j$ |

**Proposition 2:** $\hat{Z}$ that are generated by **Algorithm 2** satisfy Constraint sets (5) – (14) in **Program 1**.

**Proof:** The outer *for* loop iterates through each of the examples in our **Train**, **Test**, or **Test**$_{new}$ dataset, which corresponds to an optimization problem. Lines 1, 2 and 3 in the *for* loop ensures that Constraint sets (5) and (6) are satisfied, and the last line ensures that Constraints set (7) and (8) are satisfied. Note that we have $\hat{Z}_{0j}^{exp} = 1$ and $\hat{Z}_{i(n+1)}^{exp} = 1$ where $j$ is the node selected by the first line in the first iteration in the while loop, while $i$ is the last node selected when the while loop terminates. Constraints set (9) and (10) are ensured by iteratively selecting one, and only one $j$ throughout the iterations in the *while* loop and assign $\hat{Z}_{ij}^{exp} = 1$; the $j$ then becomes the next $i$ and the next $j$ is then selected similarly until all node in $S^{exp}$ are visited. Therefore, $\forall 1 \leq i \leq n$, we have at iteration $k$, if $i$ in $S^{exp} = \{i: \hat{Y}_i^{exp} = 1 \text{ for all } i \leq n\}$, $\sum_j Z_{ij} = Z_{ij} = 1 = Y_i$, and in the previous iteration, $\sum_j Z_{ji} = Z_{i^{k-1}j^{k-1}} = Z_{i^{k-1}i} = 1$, where $i^{k-1}$ and $j^{k-1}$ are the previous $i$ and $j$. On the other hand, if $i \notin S^{exp}$, $\sum_j Z_{ij} = \sum_j Z_{ji} = Y_i = 0$. Recall that Constraint sets (11) – (13) is the MTZ constraints that ensure that no subtour is present in the solution, where a subtour is defined as a (0)-(n+1) path that does not visit all of the open facilities in $S^{exp}$. Note that according to the definition of $S^{exp}$, node (n+1) can never be inside $S^{exp}$. Thus, the *while* loop will never set $\hat{Z}_{i(n+1)}^{exp} = 1$ if there exists another node in $S^{exp}$. Constraint set (14) is also ensured because we set $S^{exp} = S^{exp} - \{j\}$ in previous iteration, and the next $i$ (precious $j$) is already excluded from the selection.

We conclude this section with the following result:

**Proposition 3:** $\widehat{X}$, $\widehat{Y}$, and $\widehat{Z}$ together constitute a feasible solution to the Combined Set Covering and Traveling Salesman Problem that is formulated as **Program 1**.

**Proof**: From **Proposition 1** and **Proposition 2**, Constraint sets (2) – (14) are satisfied, and Constraint sets (15) – (18) are satisfied because $\widehat{X}$, $\widehat{Y}$, and $\widehat{Z}$ only contains binary values. Note that the convenience variable $U_i$ in **Program 1** is not given here and is unnecessary.

### 3.5 Evaluating the end-to-end mechanism

We use the overall cost that is defined by the objective function (1) in **Program 1** as the basis for evaluating the performance of our end-to-end ML-based mechanism to solve the Combined SCP and TSP problem. In the next section, we provide a detailed demonstration of how our approach performs by using a case study that is based upon the World Health Organization's vaccine distribution chain, and provide numerical results with data derived from four countries in sub-Saharan Africa.

## 4. An illustration: mobile clinics and outreach operations

Vaccination has been proven to be the most effective method to prevent infectious diseases. However, vaccine distribution chains can be extremely complex in many low and middle-income countries with geographically dispersed and nomadic populations, and there are thus still almost 20 million children in low and middle-income countries who remain at risk and are not covered by routine vaccines [50]. To address this problem, the World Health Organization (WHO) established the Expanded Programme on Immunization (EPI) in 1974 to provide universal access to childhood vaccines for all children. This program has successfully contributed to saving millions of lives worldwide [51]. Several papers have addressed the network design phase of the WHO vaccine distribution chains via mathematical programming [41], [52]–[55] where vaccine is assumed to be delivered to clinics via a fixed vaccine distribution chain. However, residents in remote locations in these countries often have no (or limited) direct access to clinics and hospitals. In these cases, *outreach* is typically utilized to raise immunization rates. A set of these remote population centers are chosen for locating *mobile clinics* and a team of clinicians and support personnel set up these mobile clinics periodically to vaccinate people in the immediate surrounding area.

There are a limited number of mathematical programming models to help determine optimal outreach strategies [56], [57], but none of these models looks at the problem on an ongoing basis even though outreach in practice is done at regular intervals of time, and the underlying mathematical models are required to be solved repeatedly because the same plan is not followed each time. Yang and Rajgopal were the first to present quantitative models that consider updated model parameters and obtains revised plans for subsequent planning periods using a two-period Robust approach [58]. The authors present a method to economically plan for outreach and provide management insights on where to focus more attention, but when solving the mathematical model, it starts from scratch every time that the MIP needs to be solved. Thus, there are no mechanisms to learn from historical optimization solutions..

## 4.1 Problem definition

While outreach has been proven to be an effective way to increase vaccination coverage rates in resource-deprived regions, there is no standard structure or process for outreach across all countries. In practice, a typical way that outreach is done is that a medical team departs from an existing clinic at a district center in a truck or van, while carrying vaccines in cold boxes along with related supplies. The team then sets up at one or more mobile clinic location(s) to vaccinate residents in that area as well as residents from nearby areas. Per WHO's guideline, a mobile clinic should be able to cover population centers that are within 5km of its location. In the outreach trip, if multiple locations are visited, the team goes to each of the locations sequentially and eventually returns to the original depot. Although there are often significant variations in economy, geography, demography, and thus no standard vaccine regimens and outreach practice that are identical across countries, this study aims to provide a relatively rigorous process for outreach trips across all countries to meet the WHO's goal of providing universal access to the opportunity to be vaccinated.

We take three sets of decisions into consideration: 1) choosing locations of mobile clinics as a subset of all targeted population centers to be covered; 2) assigning population centers to mobile clinics that are within the maximum coverage distance to that mobile clinic (each mobile clinic could serve multiple population centers, but a population center can only be assigned to one mobile clinic); and 3) vehicle trips that ensure that all mobile clinic locations are visited once and only once within some suitable planning horizon (e.g., 1 month). Within each planning horizon, only one vehicle trip is assumed to be undertaken and the vehicle must depart from a fixed depot and return to that depot after it visits all mobile clinic locations. The vehicles utilized in outreach trips are typically trucks or vans with several coolers or cold boxes, and since our target populations are in remote and sparsely populated area, we assume that these vehicles are not capacitated in terms of how much vaccine can be carried. Therefore, we assume that each trip could carry necessary clinical and support personnel along with the sufficient amount of vaccine for the location(s) that are served by the trip. In cases that the required vaccine volume is larger than the capacity, a larger vehicle is acquired or a shorter planning horizon can be leveraged to conduct more outreach trip throughout the year.

Three components of costs are considered in planning the outreach operation. First, we consider direct cost associated with running a mobile clinic at a particular location that includes the setup at the outreach site, labor costs for vaccination operations onsite, the cost of renting or obtaining space and storage devices, energy consumptions cost, and any other local cost. We also consider the cost of assigning population centers to a mobile clinic. This cost includes the cost of moving targeted newborns from other population centers without a mobile clinic, incentives paid to them to have them visit a mobile clinic, and estimated social and healthcare cost associated with people not visiting mobile clinics due to relative long travel distances. The assignment cost can be formulated as a linear function of distance from population centers to mobile clinic, although this is not crucial. Lastly, we consider trip-related cost that includes vehicle depreciation or vehicle rental costs, fuel costs, hourly wages/allowances paid to the team and driver. Note that this cost is assumed to be proportional to the duration of the trip, and we thus utilize an average cost per hour based on the trip duration to quantify it. In summary, the total cost is determined by the selected mobile clinics locations, distance from population centers to mobile clinics, and the route taken by the vehicle on the outreach trip.

This process can be viewed as an example of a combination of the Set Covering Problem and the Traveling Salesman Problem as discussed in Section 1. The selection of mobile clinics can be viewed as the selection of facility locations in $Y$, assigning population centers to the selected mobile clinics can be considered via assignment variables $X$, and the route to visit mobile clinics can be viewed as the TSP route $Z$. Moreover, because the demand, traffic and road conditions in these targeted zones are typically unstable, it can be challenging to obtain an accurate set of estimates for these problem parameters ahead of time. For example, the demand at a population center is calculated by its population and the prevalent birth rate. Thinking of demand as being stochastic can sometimes be more accurate, as both the population and the birth rate within a location could vary every year, or even within a particular year because of seasonal movements of the population. Similarly, travel time and assignment cost from $i$ to $j$ are often not constant because road and traffic conditions in these targeted zones are often unstable. With extreme events such as epidemics, landslides or floods, traffic can be impacted or even blocked. On the other hand, improvements to infrastructure can also reduce travel times and assignment costs. Therefore, it would be ideal to determine a flexible strategy over every successive planning period. The outreach model thus needs to be solved periodically with similar parameter and inputs, where each instance corresponds to a graph with one depot and multiple population centers. The learning-based mechanism described in Section 3 therefore constitutes a promising direction from which to approach this problem, and we thus apply the methodology illustrated in Section 3 to solve this problem and present the numerical results using data derived from the WHO and four countries in sub-Saharan Africa.

## 4.2 Numerical results

To present numerical results using the mobile clinic and outreach operations example, we selected 30 unique sets of targeted populations across different parts of four countries in sub-Saharan Africa, and for each such set, we generated the node set of locations and 1,000 different examples using different combinations of cost data. This generated a total of 30,000 different instances. We ran these 30,000 examples on **Program 1** using Gurobi over a period of several weeks, and obtained the optimal solution vector ($X^*$, $Y^*$, $Z^*$) for each instance. We also documented each cost component (facility location cost, assignment cost, and TSP cost) in each of the optimal solutions.

We then randomly picked 5 distinct sets of population centers from the 30 we started with and assigned all of the corresponding 5,000 examples associated with the 5 graphs in this set to dataset **Test$_{new}$**. For the remaining 25 population centers and the corresponding 25,000 instances, we randomly split these into 22,500 examples for dataset **Train** and 2,500 examples for dataset **Test**. Note that instances in **Train** and **Test** are drawn from the same pool of locations, so that the same population center and its associated graph could appear in both datasets, albeit with different cost information. However, the graphs in **Test$_{new}$** are all different from those in **Train** and **Test**. In other words, the instances in **Train** and **Test** bear some resemblance to each other, unlike **Test$_{new}$**, whose instances come from a completely different set of population centers and their graphs. After utilizing the method introduced in Section 3.2 and 3.3 to train the two machine learning models on **Train** with the data set split into training and validation sets and parameter tuning after each iteration, we obtain the final model parameters. Given any instance, we then use machine learning **Model 1** to obtain $g(Y)$ followed by **Algorithm 1** in Section 3.4 to obtain SCP solution $\widehat{X}$ and $\widehat{Y}$. We then feed $\widehat{X}$ and $\widehat{Y}$ into machine learning **Model 2** to obtain $p(Z)$ followed by **Algorithm 2** to obtain TSP solution $\widehat{Z}$.

We use the datasets **Test** and **Test$_{new}$** to evaluate how well the machine learning models (along with the two algorithms for obtaining a final solution) perform on data that is distinct from the data used to train. To evaluate the end-to-end performance of the approach, we compare the total cost of the solution obtained with the minimum cost obtained by solving Program 1. We started with a value of 0.5 for the threshold $\alpha$ described in Section 3.4 since this would appear to be a natural value for it if we interpret each element of $g(Y)$ as the probability that the corresponding variable is equal to 1. We summarize the numerical results for each component of cost in Table 3. For each component and data set, the entry in Table 3 is the ratio of the sum of the costs for that component across all instances in the data set obtained from our approach, and the sum of the optimal costs for that component across the same instances obtained by solving the MIP (expressed as a percentage).

Table 3. Cost comparison of each component in **Train**, **Test** and **Test**$_{new}$

|              | # examples | Facility cost | Assignment cost | TSP cost | Total Cost |
|--------------|------------|---------------|-----------------|----------|------------|
| **Train**    | 22,500     | 100.09%       | 100.25%         | 102.02%  | 101.20%    |
| **Test**     | 2,500      | 100.34%       | 100.42%         | 102.36%  | 101.49%    |
| **Test**$_{new}$ | 5,000  | 147.56%       | 58.77%          | 158.80%  | 146.33%    |

Since examples in **Train** and **Test** are drawn from a common pool of population centers, while examples in **Test**$_{new}$ are from a completely different set of population centers, it is natural that the approach will yield better results with instances in **Test** than with instances in **Test**$_{new}$. The numerical results from data set **Test** should give us a fair measurement of the performance of the mechanism on future examples from node sets that have been solved before albeit with different cost parameters. On the other hand, data set **Test**$_{new}$ presents a more "tough" test to measure the model's generality, because these instances use node sets that were never part of the training process and thus no actual information on historical solutions is given to the learning mechanism.

As shown in Table 3, with instances in **Train** and **Test** our approach is able to generate heuristic solutions that are on average about 1% more expensive than the optimal solution from solving **Program 1**. Because the mechanism utilized a train-validation splitting procedure within the backend training and hyperparameter tuning process, overfitting is not observed, and the mechanism preforms similarly on both data sets.

On the other hand, when dealing with optimization problems with a different pool of population centers that are not part of the training process, the model is not able to generate solutions that are as good. In particular, it appears that the machine learning approach performs notably worse on the **Test$_{new}$** set, and leads to facility costs that are 47.56% higher than the optimal cost. When it comes to new datasets that it has never seen, **Model 1** is not as good at identifying the correct combination of facility locations to cover all population centers, and consequently **Algorithm 1** turn out to be more "conservative" and opens a number of additional locations to ensure complete coverage, leading to higher facility costs. However, since more mobile clinics are selected, the corresponding total assignment costs are only 41.23% of what they are at the optimum, because on average, with more clinic locations there are fewer population centers assigned to each clinic and population centers now get assigned to clinics that are closer. More clinics also lead to outreach trip costs that are 58.80% higher, since there are now more stops to visit in the final solution. Overall, the cost is around 46% higher than the optimal total cost on average. Note that in this study, it is more common to have higher facility and transportation cost, because vaccine has to be transported and stored in a very narrow range of temperature, and often times special types of storage devices and vehicles have to be used. In the optimal solution, **Program 1** thus tries to select fewer clinic locations, with each serving more population centers, as opposed to having more clinic locations serving fewer population centers. If the mechanism would be used in a scenario with more balanced cost components, we might expect it to perform better with the **Test$_{new}$** data set.

One observation in solving via this mechanism that needs to be mentioned is run time. Although the training procedure can take several hours with the 22,500 observations, when it actually comes to predicting probabilities and translating these into a feasible solution for a specific instance, the time is negligible. On the other hand, solving the problem via **Program 1** can take much more time, and in applications where an immediate solution is desirable, the proposed mechanism is more favorable over **Program 1** once training has been completed. Overall, the mechanism is able to generate high quality results repeatedly for problems that resemble instances in the training set, but when it encounters totally new problems, it generates more expensive heuristic solutions but that are guaranteed to be feasible.

### 4.3 Selecting a value for the parameter $\alpha$

In Section 4.2 we discussed the numerical results when parameter $\alpha$ is set to a natural value of 0.5 and is consistent throughout the implementation of **Algorithm 1** on **Train**, **Test** and **Test$_{new}$**. However, as illustrated in Table 3, the total cost of our solutions on **Test$_{new}$** is significantly higher than the total cost based on the optimal solution. In particular, the facility cost in our solutions is higher than the optimum, and more facilities then lead to longer TSP routes, which in turn are also higher than the optimum, and these together overwhelm the reductions in assignment costs. To further study parameter $\alpha$ and its impact on the total cost, we conducted a sensitivity analysis on parameter $\alpha$. Table 4 reports each component of cost across instances in **Test$_{new}$**, similar to what we had in Table 3 for $\alpha = 0.5$, but with different values of parameter $\alpha$ from 0.1 to 0.9 in **Algorithm 1**. Note that $\alpha$ is set to be consistent throughout each of the scenarios from $\alpha = 0.1$ to $\alpha = 0.9$.

**Table 4**. Sensitivity analysis of parameter $\alpha$ on **Test$_{new}$**

| $\alpha$ | Facility cost | Assignment cost | TSP cost | Total Cost |
|---|---|---|---|---|
| 0.1 | 158.62% | 49.62% | 168.46% | 154.92% |
| 0.2 | 154.35% | 52.90% | 164.60% | 151.51% |
| 0.3 | 151.94% | 54.97% | 162.58% | 149.69% |
| 0.4 | 149.64% | 56.96% | 160.50% | 147.88% |
| 0.5 | 147.56% | 58.77% | 158.80% | 146.33% |
| 0.6 | 145.72% | 60.52% | 157.14% | 144.90% |
| 0.7 | 143.97% | 62.24% | 155.67% | 143.59% |
| 0.8 | 141.60% | 64.63% | 153.61% | 141.79% |
| 0.9 | 138.48% | 67.75% | 151.08% | 139.52% |

It can be seen that as $\alpha$ increases from 0.1 to 0.9, the predicted total cost in **Test$_{new}$** improves from being 54.92% higher to being 39.52% higher than the optimal solution, with a decrease in facility cost and TSP cost from 58.62% and 68.46% higher to 38.48% and 51.08% higher, respectively. This indicates that **Algorithm 1** appears to perform better with higher $\alpha$ values when the initial number of facility locations is relatively small. One possible explanation for this is that often times when two population centers are close to each other and also resemble each other in terms of their demand, from a machine learning perspective, **Model 1** would tend to predict similar probabilities for locations at both. In situations where these probabilities are relatively large and $\alpha$ is relatively small, **Algorithm 1** yields a solution with facilities open at both locations. However, in practice we would only want a facility to be open at one of these locations; the other one could be covered by the open facility if the distance between the two is within the MCD, and many other population centers might be close enough to be served by either location. On the other hand, with a higher $\alpha$, the likelihood of this happening is smaller because there will tend to be fewer open locations overall, thus reducing the likelihood of this type of duplication.

Furthermore, when the probabilities are on either side of $\alpha$, but the difference is small (e.g.,0.89 vs. 0.91, with $\alpha$=0.9), higher values of $\alpha$ could prevent both from being opened (in our example, only one facility will be opened, whereas both would be open with $\alpha = 0.8$.

To further study this, we treated $\alpha$ as a decision variable, and ran multiple threads of the process in Section 3.4 with different values of $\alpha$ simultaneously. Here, for each specific instance in **Test_new**, we picked the value of $\alpha$ that yields the minimum total cost ($\alpha^*_{exp}$), and then calculate the corresponding cost components. In examining the values of $\alpha^*_{exp}$ that yielded the lowest total costs, we found that in 2,036 of the 5,000 instances (~41%), the choice of a value for $\alpha$ made no difference at all. This is because the majority of the probabilities in these instances are in the range (0, 0.1), or (0.9, 1.0), so that any value of $\alpha$ between 0.1 and 0.9 would give the same solution. In many other instances, there was a range of values for $\alpha^*_{exp}$ that yielded the same solution; in fact, in over 75% of the instances there were at least three (consecutive) values for $\alpha^*_{exp}$ that were optimal.

To understand this better, Figure 1 shows the count of different values of $\alpha$ that are optimal across the 5,000 instances in **Test_new**, noting again that in general we have multiple optimal of $\alpha^*_{exp}$ at the same time for almost all instances. Figure 1 also shows that our solutions improve as $\alpha$ increases and in 4,577 instances (~92%) the value of $\alpha$=0.9 yielded the best solution from using our approach. In Table 5 we further compare each individual component of cost to the optimum, similar to what we had in Table 4. The first row in Table 5 summarizes costs when we pick $\alpha = \alpha^*_{exp}$ for each instance. Overall, our results indicate that simply picking a value of $\alpha$=0.9 provides us with virtually the same result (139.52% from the last row of Table 4) as picking the best possible value for $\alpha$ (=$\alpha^*_{exp}$) for each instance (138.75% from the first row of Table 5). As a matter of interest we also show in each subsequent row of Table 5 cost comparisons from different values of $\alpha$ from 0.1 to 0.9, but only for those instances for which $\alpha = \alpha^*_{exp}$. Again, comparing the last row of Table 4 with the first and last rows of Table 5 shows that a value of 0.9 for $\alpha$ yields the best results for the vast majority of instances.

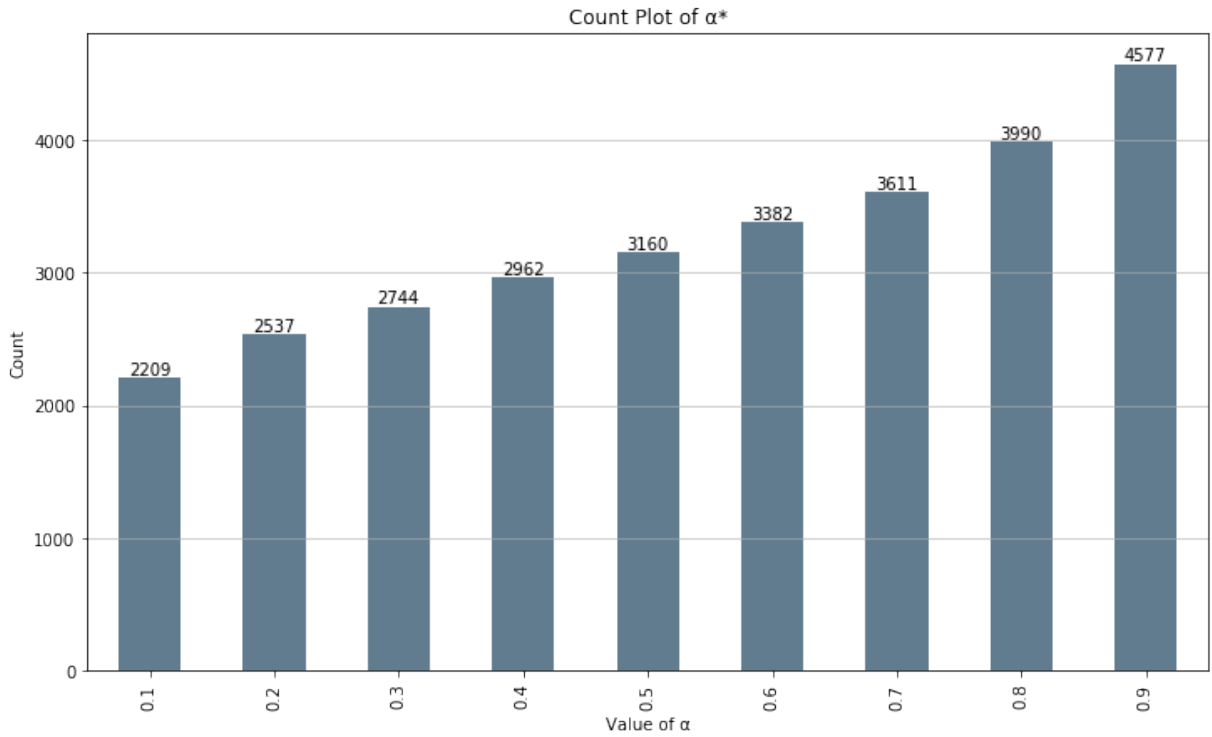

**Figure 1.** Count of examples in **Test**_new_ for which $\alpha^*_{exp}$ is best

**Table 5**. Cost comparison of each component in **Test**_new_ with different $\alpha^*_{exp}$

| $\alpha$ | Facility cost | Assignment cost | TSP cost | Total Cost |
|---|---|---|---|---|
| Overall | 138.79% | 66.82% | 149.63% | 138.75% |
| 0.1 | 143.48% | 58.94% | 155.68% | 143.79% |
| 0.2 | 142.96% | 59.64% | 155.15% | 143.20% |
| 0.3 | 142.77% | 60.55% | 154.76% | 142.87% |
| 0.4 | 142.61% | 61.08% | 154.31% | 142.56% |
| 0.5 | 142.17% | 61.83% | 153.86% | 142.16% |
| 0.6 | 141.58% | 62.54% | 153.19% | 141.67% |
| 0.7 | 140.75% | 63.67% | 152.47% | 141.00% |
| 0.8 | 139.94% | 64.99% | 151.73% | 140.30% |
| 0.9 | 138.80% | 66.86% | 150.57% | 139.32% |

Note that in theory, $\alpha_{exp}^*$ could also be treat as a component of the dependent variables and can thus be modeled and predicted via a machine learning model. We will discuss this along with some other future research directions in Section 5.

## 5. Discussion

There are several limitations from this initial work that provide us with future research directions. First, as discussed at the end of Section 4.3, we could learn the optimal threshold parameter $\alpha_{exp}^*$ in **Algorithm 1** via machine learning, and include it as a component of the output in **Model 1**:

$$\text{Model 1}': G \rightarrow g(Y), \alpha$$

i.e., **Model 1**′ not only establishes the probability that in the optimum solution, the facility at location $j$ is open, but also the threshold to consider opening the facility at location $j$ in **Algorithm 1**. Note that the training process of **Model 1**′ must be iterative, because the $\alpha^*$ is unknown until the different threads of the whole mechanism including **Model 2** is completed in parallel.

Second, as discussed in Section 4.3, **Model 1** tends to predict similar probability when two nearby population centers resemble each other. With a small $\alpha$, we could end up with both having open facilities, and this results in much higher facility costs then necessary. This leads to another promising direction in the future to utilize techniques such as Convolutional Neural Network (CNN) to ensure that the Set Covering model considers the correlation between population centers.

Third, there could be a mechanism with feedback between the models, as opposed to training each independently. As a new direction, we could leverage the information that is available within each machine learning model to help with improvement across both models. For example, the TSP cost that is calculated from **Model 2** can be viewed as a function of the output of **Model 1 (**which is the SCP solution). If the TSP cost that is incurred by a particular SCP solution can be utilized as intermediate information when training **Model 1**, it is possible that **Model 1** could generate better SCP solutions that leads to smaller overall cost. One way of doing this might be to initially use an approximation of the TSP cost to help speed up the learning process. Overall, this direction requires a more sophisticated methodology to implement.

Fourth, in Section 4, because all examples considered in the demonstration have a similar number of population centers (around 10), we utilized a standard, fully connected neural network in the implementation of **Model 1** and **Model 2**. This places some limitations on the generality of the mechanism and the possibility of utilizing the mechanism to solve problems with potentially more nodes. In many real-world circumstances, the number of facility locations could vary over a wider range and could be unknown before the start of the training process. This points to the possibility of using more general, sequential neural networks whose dimension of input is not fixed (unlike with the standard version we used). For example, using Recurrent Neural Networks (RNN) such as Long Short Term Memory (LSTM) networks could potentially alleviate this problem. In developing RNN, we can use a structure similar to LSTM that is widely used in Natural Langrage Processing and time-series analysis. Information that is related to a particular node in our graph can be fed into the RNN sequentially, and the predicted probabilities related to this node along with the information of the next node are then fed into the next unit of the RNN, until the information across all nodes is fed into the model and all prediction results are obtained.

Fifth, in this work the training procedure to establish the machine learning models is conducted on the entire training dataset before the prediction process. However, in most situations, especially when inputs are coming in as a flow or when optimal solutions may change dynamically over time because of changes in parameter values, the machine learning models also need to be updated over time. One natural way is to retrain the models on the entire training set over time after a given set of planning horizon, but this could be time consuming. Alternatively, we could define a preselected number of inputs as a "batch" and once we receive a whole batch of new input, we update the parameters of the machine learning models sequentially by batch. This technique is often referred to as online learning, or incremental learning, and is widely used when near real-time inputs are present.

Lastly, from an implementation standpoint, we would like to seek more real-world circumstances that are applicable for the mechanism. There are often two challenges. First, to obtain mechanisms that are more consistent and general, it is necessary to have a different number of nodes, and a wider range of different parameter distributions. The historical optimization solution should also be available to train the machine learning model. Secondly, there must be a need to have the optimization model solved repeatedly and rapidly.

# 6. Summary

This paper aims to provide an early explorations and experiments in leveraging a machine learning algorithm to solve a difficult combinatorial optimization problem. We first study the combined Set

Covering and Traveling Salesmen problem, and formulate this problem as a Mixed Integer Program. When the optimization problem needs to be solved on a regular basis with similar input values, it starts from scratch as new inputs and parameters are given. This could lead to high computational expense as well as tractability issues. To address this, we introduce a machine learning based mechanism to solve this problem by leveraging historical optimal solutions. The mechanism can be utilized to efficiently generate heuristic solutions via two machine learning models that are dedicated to solving the Set Covering Problem and the Traveling Salesmen Problem separately, but are aimed at minimizing overall cost. We discuss data generation and preprocessing, model training, and how to generate feasible solutions using the machine learning results. We also discuss how to compare the overall cost of the machine learning based mechanism to the optimal cost that is generated by the optimization formulation. We then present a detailed case study for the World Health Organization's vaccine distribution chain, and provide numerical results with data derived from four countries in sub-Saharan Africa after several train-test-evaluation iterations. Based on the computational performance observed, the machine learning based mechanism appears to be a promising way to achieve tractability and scalability without significantly sacrificing solution quality, but it still requires significant further development and should supplement the current exploratory approaches of incorporating machine learning with optimization.

**Acknowledgement**

This work was partially supported by the National Science Foundation via Award No. CMII-1536430.